# Multilingual Models for Check-Worthy Social Media Posts Detection

Sebastian Kula[1,2]

[1] Kempelen Institute of Intelligent Technologies, Bratislava, Slovakia
[2] West Pomeranian University of Technology in Szczecin, Szczecin, Poland

**Abstract.** This work presents an extensive study of transformer-based NLP models application for detection of social media posts that contain verifiable factual claims and harmful claims. The study covers various activities, including dataset collection, dataset pre-processing, architecture selection, setup of settings, model training (fine-tuning), model testing, and implementation. The study includes a comprehensive analysis of different models, with a special focus on multilingual models where the same model is capable of processing social media posts in both English and in low-resource languages such as Arabic, Bulgarian, Dutch, Polish, Czech, Slovak. The results obtained from the study were validated against state-of-the-art models, and the comparison demonstrated the robustness of the proposed models. The novelty of this work lies in the development of multi-label multilingual classification models that can simultaneously detect harmful posts and posts that contain verifiable factual claims in an efficient way.

## 1 Introduction

The application of NLP methods, spurred by the development of transformer architectures, has expanded to various subject domains. Nowadays, many works focus on exploring the possibility of deploying NLP techniques to combat misinformation, fake news, or propaganda [12]. Although there seems to be a consensus in the community that completely replacing humans in the process of detecting fake news is currently not feasible, transformer architectures are being studied as solutions that can significantly optimize and improve the work of human fact-checkers [27]. With the development of electronic communication, particularly social media, and the simultaneous increase in awareness and responsibility of governmental, social, and opinion-forming institutions, the need to create applications and tools to combat fake news and disinformation is growing [18].

The ambiguity and diversity of natural languages, the high intensity of irony and satire in social media texts and posts, and the presence of cultural context make the task of detecting fake news a challenging and time-consuming process [27]. The process requires the involvement of fact-checkers, i.e., experts with specific knowledge in each particular domain. To relieve overwhelmed fact-checkers and facilitate their work, NLP and DL (deep learning) methods are being applied. This paper focuses on AI models that facilitate the work of human fact-checkers and presents models for the detection of verifiable factual claims (vfc) and harmful claims.

Detecting check-worthy claims is the first step in the process of detecting fake news [4]. Verifiable factual claims are posts that state a definition, mention a quantity in the present or the past, make a verifiable prediction about the future, reference laws, procedures, or rules of operation, discuss images or videos, and make statements regarding correlation or causation [22]. Harmful claims are defined as offensive and/or hateful content on social media that can harm individuals, organizations, and the society [22].

### 1.1 Motivation

The main motivation of the research was to facilitate the work of fact-checkers by creating an applicable NLP tool that detects simultaneously verifiable factual claims and harmful claims. Based on the discussions with engineers supporting human fact-checkers this tool should be easy to implement, not overly demanding in terms of hardware, and at the same time applicable to CPU units and standard GPU units found in cost-effective computing devices. The tool should have the ability to infer thousand of posts in a maximum of one hour, so that fact-checkers can verify facts on an ongoing basis. The tool should perform well for low-resource languages like Arabic, Bulgarian, Dutch and especially Polish, Czech and Slovak.

In order to meet above-mentioned requirements, the paper addresses the following scientific questions and analyses the following technical challenges: a) whether it is better to use translations and models dedicated to English or whether it is better to use multilingual models for the task of verifiable factual claims detection; b) whether multilingual and multi-label models perform correctly for low-resource languages in the verifiable factual claims detection task; c) whether a multi-label model that simultaneously detects verifiable factual claims and harmful claims performs correctly, and whether such a single model gives sufficiently high results; d) whether there are linguistic cues to facilitate the automatic detection of verifiable factual claims and harmful claims. The answers to these research questions make it possible to identify an applicable NLP architecture, that is efficient in terms of hardware, efficient in economic terms, with a relatively low inference time and simultaneously maintaining high performance of results.

### 1.2 Outline

This paper presents a series of experiments that involve training (fine-tuning) existing transformer architectures to perform check-worthy claim detection tasks, which are also classification tasks. The following architectures were fine-tuned: DistilBERT [30], BERT-large [6], BERT-base [6], XLM-RoBERTa-base [5], and XLM-RoBERTa-large [5]. The study utilized the Flair tool [3] and the cloud computing platform Google Colaboratory [11]. An important aspect of this work was to create models that would not only be characterized by a high level of performance but that would also be easy to implement, use, and maintain. In addition to analyzing typical NLP model metrics for classification tasks (such as accuracy, recall, and F1-score), inference times were also analyzed.

The paper presents the entire pipeline starting from data selection and preprocessing, through architecture selection and fine-tuning, to model testing, and additional validation in real-life applications. Three types of models were obtained: models for detecting verifiable factual claims, models for detecting harmful claims, and models that simultaneously detect both types of claims. The proposed solutions are reliable and useful, as confirmed by the verification of the results. The training data used in the pipeline covers multiple subjects, making the models effective at detecting check-worthy claims in posts on multiple topics.

### 1.3 Key Contributions

The key contributions of the paper are as follows:

- Creation of models based on XLM-RoBERTa-base architecture for simultaneous detection of verifiable factual claims and harmful claims in multilingual posts. The models were tested for the following languages: English, Turkish and low-resource languages such as Arabic, Bulgarian, Dutch and also Polish, Slovak, and Czech.
- Selection and composition of multi-subject and multilingual data collections containing verifiable factual claims, which allowed for the creation of the aforementioned models.
- Conducting a detailed inference time analysis of the models implemented using (a) only CPU units and (b) also GPU units. The obtained times show that the models' inference is much faster than that of humans.

## 2 Related Works

With the emergence of transformer architectures in the NLP domain, automated fake news detection solutions have become increasingly prevalent. In [19], the BERT architecture [6] and other derivative architectures such as RoBERTa [21], DistilBERT [30], autoregressive DistilGPT2 [30], and XLNet [35] were applied to classify news articles, mainly on political and social topics, into articles that contain or do not contain fake news. Transformer-based models were also applied in the fight against COVID-19 misinformation in [10,20,17], where they achieved state-of-the-art results.

The task of fact-checking is also gaining prominence in research [12]. The first step in the fact-checking pipeline is detecting claims in posts. Fact-checking methods not only determine whether a post is true or false (binary classification), but also verify and potentially debunk disinformation contained in it. This can be achieved e.g. by highlighting verifiable claims in the text and linking to relevant sources for fact-checkers.

Several articles discuss the task of claim detection using various transformer architectures, including [13,33,29,22]. Stammbach et al. [33] applied fine-tuned DistilBERT [30], RoBERTa [21], and ClimateBERT [34] to detect environmental claims using a dataset designed specifically for the environmental domain. The CLEF2022 competition [22] focused on the detection of verifiable factual claims and harmful claims, with the best results achieved by methods based on BERT [9], XLM-RoBERTa [31], and GPT-3 [2]. Authors of [14] used a claim-spotting system, called claimBuster to detect sentences containing claims in news articles about COVID-19. [13] presented models for detecting claims in any type of online text.

## 3 Overview of the Models' Architectures

To establish an optimal model for detecting social media posts with verifiable factual claims and harmful claims, we conducted a number of experiments using architectures based on Transformers, namely DistilBERT, BERT-large, BERT-base, XLM-RoBERTa-base, and XLM-RoBERTa-large. We applied standard versions of these architectures and made no additional changes to their hyperparameters, such as the typical number of layers or the number of self-attention heads. We trained the models as classifiers, with the last (top) layer in each model being a linear dense layer.

After conducting experiments and analyzing the requirements for the claim detection tasks, we chose the multilingual XLM-RoBERTa architecture to obtain the final models, due to the multilingual nature of the detection task. The decision to select XLM-RoBERTa architecture was also justified by a literature review [7], [25], [22]. See Appendix A in additional materials `https://github.com/SebastianKula-ai/multi4CWSocial`.

The layers of the XLM-RoBERTa-base architecture are presented in the Appendix A.

## 4 Datasets

Due to the very specific and unique tasks considered here, the availability of appropriate labeled data was very limited. The following datasets were used in this work: CLEF2022 (task 1B and 1C) [22], CLEF2021 (task 1B) [32], LESA2021 (noisy and semi-noisy datasets) [13], and the MultiClaim dataset [26].

Dataset CLEF2022 (task 1B) contains Twitter posts dedicated to the topic of COVID-19, which are binary-labeled as containing or not containing verifiable factual claims. The dataset contains five different languages (English, Turkish, Dutch, Arabic, Bulgarian) and is highly unbalanced, with many more items labeled as 1 (containing verifiable factual claims) than 0 (without verifiable factual claims). Table 1 shows the number of posts in each category across the included languages.

Dataset CLEF2022 (task 1C) is very similar to CLEF2022 (task 1B) – it contains posts almost identical to those in CLEF2022 (task 1B), but labeled for containing or not containing posts with harmful claims. This dataset is also unbalanced, with significantly more 0 (not harmful) posts than 1 (harmful) posts.

Dataset CLEF2021 (task 1B) contains data on political debates, and the content of the collection refers to a variety of topics that are the subject of current political debates. The samples were labeled as containing or not containing fact-check-worthy verifiable factual claims. The dataset is characterized by a large imbalance, with more non-check-worthy elements that do not contain verifiable factual claims.

The LESA2021 dataset is a collection of data on a variety of topics collected from Twitter (COVID-19 topic) and six publicly available benchmark datasets [13].

Finally, the MultiClaim dataset is a unique collection of data containing posts and verified factual claims paired with them. The topics of the posts are diverse, and the collection is also multilingual.

The presented datasets were used to create several collections. An overview of these collections is given in Table 2. Each collection was split into train/validation/test folds (using [3]). The sizes of the splits are shown in Appendix B of additional materials.

### 4.1 Collections

Next we are going to describe each individual collection in more detail.

*Collections 1 & 1tr* Collection 1 is based entirely on the CLEF2022, task 1B dataset and contains all items and all five original languages. The task is to classify posts as containing or not containing verifiable factual claims. There is another version of this collection, called collection 1tr, in which all non-English (i.e., Turkish, Dutch, Arabic, and Bulgarian) posts were automatically translated into English using Google Translate. The collection contains the total of 22,867 posts.

*Collections 2 & 2tr* Collection 2 is based entirely on the CLEF2022, task 1C dataset and the original five languages. The task is to classify posts as harmful or not. A version of this collection with only English posts was also created, which we refer to as collection 2tr. The collection contains the total of 22,867 posts.

| Dataset language | Nr. of posts with verifiable factual claims | Nr. of posts without verifiable factual claims |
|---|---|---|
| English (EN) | 3,040 | 1,753 |
| Turkish (TR) | 2,480 | 1,331 |
| Dutch (NL) | 1,861 | 2,162 |
| Arabic (AR) | 4,121 | 2,093 |
| Bulgarian (BG) | 2,697 | 1,329 |
| **Total** | **14,199** | **8,668** |

**Table 1.** Amount of posts with and without verifiable factual claims in the CLEF2022 (task 1B) dataset [22].

| Coll. # | Datasets | Languages | Claim Detection Task | In |
|---|---|---|---|---|
| 1 | CLEF2022:1B | EN, TR, NL, AR, BG | vfc | exp2 |
| 1tr | CLEF2022:1B | EN + (TR, NL, AR, BG) → EN | vfc | exp1 |
| 2 | CLEF2022:1C | EN, TR, NL, AR, BG | harmful | exp3 |
| 2tr | CLEF2022:1C | EN + (TR, NL, AR, BG) → EN | harmful | exp4 |
| 3 | CLEF2022:1B; Coll. 2; Subset of: MultiClaim; CLEF2021:1B; LESA2021 | EN, TR, NL, AR, BG + CLEF2022:1B → SK, CS, PL | vfc + harmful | exp5 |

**Table 2.** Collections and data splits used in the experiments.

*Collection 3* Collection 3 is a combination of CLEF2022, task 1B with selected data from the MultiClaim dataset [26] and with selected data from CLEF2021, task 1B. Concerning the MultiClaim dataset, the selection of elements for collection 3 was based on the selection of posts scripted in the Latin alphabet or in Bulgarian or Arabic language. All posts selected from the MultiClaim dataset were labeled as 1 (containing verifiable factual claims). The selection of posts from the CLEF2021, task 1B dataset consisted of selecting only posts labeled as 0 (posts without verifiable factual claims).

Collection 3 also aggregated posts from CLEF2021, task 1B by combining three posts into one post. The purpose of this operation was to avoid bias related to the length of posts. CLEF2021, task 1B contains very short posts relative to other items in collection 3, hence the requirement of the aggregation.

Collection 3 also contains selected semi-noisy items from LESA2021 [13], labeled as 0 (posts without verifiable factual claims), along with posts translated from the CLEF2022 (task 1B) collection into Slovak, Czech, and Polish. Finally, collection 3 also includes the entire collection 2 with items labeled as harmful vs. non-harmful posts. In summary, collection 3 contains four labels (verifiable factual claims, non-verifiable factual claims, harmful posts, non-harmful posts) and the total of 63,118 posts.

### 4.2 Pre-Processing

All described collections have been pre-processed in accordance with the requirements of the conducted experiments. Two methods of pre-processing were applied. Pre-processing method 1 consisted of eliminating punctuation, URLs, e-mail addresses, white spaces, emoticons, newlines, and empty lines. The motivation behind this was to eliminate aspects of the posts that are unlikely to be relevant to the detection of verifiable factual claims and harmful content and are prone to act as noise in this context.

| Exp. # | Task | Languages | Collection | Architectures |
|---|---|---|---|---|
| 1 | claim/non-claim | Translated | 1tr | BERT-large-uncased |
| 2 | claim/non-claim | Multilingual | 1 | XLM-RoBERTa-base/large<br>multilingual DistilBERT-base<br>multilingual BERT-base |
| 3 | harmful/non-harmful | Multilingual | 2 | XLM-RoBERTa-base/large |
| 4 | harmful/non-harmful | Translated | 2tr | DistilBERT-base |
| 5 | claim/non-claim +<br>harmful/non-harmful | Multilingual | 3 | multi-label<br>XLM-RoBERTa-base |

**Table 3.** Overview of the experiments (type of task, data collection, trained architectures).

Social media posts tend to contain a lot of incorrect punctuation, which could impact the performance of the models negatively. While human-readable URLs could potentially contain generalizable content, the URLs in the dataset were generally either shortened URLs or links to other social media posts that differed just by their numeric identifiers. Finally, emoticons are often used in social media posts to express emotions, but they generally do not convey any factual information. Elimination of emoticons in the check-worthy claim detection task was proposed in paper [1] and URLs in paper [15].

Pre-processing method 2 extends method 1 by also eliminating posts that are not scripted in Bulgarian, Arabic, or Latin alphabet, this is due to the MultiClaim dataset, which contains alphabets outside the scope of this work. More details about pre-processing are presented in Appendix C of additional materials.

## 5 Experimental Setup and Execution

| Hyperparameter name | Value |
|---|---|
| learning rate | 3e-05 |
| batch size | 32 |
| anneal factor | 0.5 |
| patience | 3 |
| max number of epochs | 5/10*/15** |
| mini batch chunk size | 1*** |

**Table 4.** Training hyperparameters' values; * 10 epochs were applied for XLM-RoBERTa-large;** 15 epochs were applied for DistilBERT-base; ***mini batch chunk size was set only for XLM-RoBERTa-large due to GPU memory limitations.

The data preparation process and the execution of the experiments were performed remotely on the Google Colaboratory cloud platform [11], using the Flair tool (version 0.7) [3]. The experiments were conducted on an instance with a Tesla T4 card with 16 GB RAM and an Intel Xeon CPU @2.00 GHz with 12.68 GB RAM.

To identify the optimal model for detecting posts with claims, we conducted five experiments. Two of them focused on detecting posts containing verifiable factual claims (experiment 1 and experiment 2), and the next two were related to detecting posts containing harmful claims (experiment 3 and experiment 4). The final experiment (experiment 5) concerned the multi-label model, which simultaneously performed the task of detecting posts containing verifiable factual claims and harmful claims.

In all experiments, the data were split into train/validation/test folds. The sizes of the folds are presented in Appendix B. The train set was used for training, and the validation (dev) set was used to determine validation accuracy at each epoch. During training, both the last model (trained for the most epochs) and the best model (in terms of validation accuracy) were checkpointed. The best models and the last models were then taken and tested on the withheld test set. Thus the number of epochs hyperparameter was tuned. The top results were presented.

Table 3 provides an overview of the experiments, including what the target task was, what data they used, and what architectures were applied. The hyperparameters for all experiments are presented in Table 4.

One thing to note about the experimental setup is the choice of architectures. Experiment 1 is our baseline experiment, where all non-English posts were machine-translated into English. For this experiment we used BERT-large-uncased to get a robust baseline to compare against.

In subsequent experiments, we opted for somewhat smaller models (which showed comparable performance), specifically we used DistilBERT-base for the cases where all non-English posts were machine-translated into English and multilingual DistilBERT, multilingual BERT-base and XLM-RoBERTa-base/large for the multilingual setup.

Experiment 5 was the most crucial one for this work and its scope and shape were the result of the other four experiments. The focus was (i) to create a single model capable of posts containing verifiable factual claims and harmful claims in a multi-label setup; (ii) create a model which can also detect posts not related to COVID-19 topics. This is why collection 3 was created.

## 6 Results and Evaluation

The models were compared in terms of recall, f1-score, and accuracy, which are the typical metrics used to evaluate NLP classifiers. The models were also compared in terms of the requirements during the implementation phase of the models in hands-on applications and in terms of the performance of the models in the inference process.

The developed models were compared with the results published in CLEF2022 [22], as well as, due to the unavailability of results in SK, CS, PL languages with the multilingual models Nithiwat xlm-roberta-base claim-detection [24] and Nithiwat mdeberta-v3-base claim-detection [23] available on Hugging Face. The final multi-label model was also compared with the Nithiwat model and LLMs (large language models) like: Alpaca-LoRA [16], LLaMa-3.1-405b-instruct [8], LLaMa-3.1-70b-instruct [8], Qwen3.5-397b-a17b [28] for the tasks of verifiable factual claims and harmful claims detection. Details of testing (including McNemar test) models from Hugging Face and Alpaca-LoRA and further comparisons are in the Appendix A.

| Model | Exp. # | Accuracy | Recall | Test dataset |
|---|---|---|---|---|
| XLM-RoBERTa-large 10 epochs | 2 | **0.7558** | 0.7724 | 5 original languages |
| BERT-large-uncased | 1 | 0.7388 | **0.7938** | EN + translations |
| XLM-RoBERTa-base | 2 | 0.7520 | 0.6976 | 5 original languages |
| distilBERT-base-multilingual-cased | 2 | 0.7223 | 0.6811 | 5 original languages |
| BERT-base-multilingual-cased | 2 | 0.7239 | 0.6597 | 5 original languages |

**Table 5.** Comparison between fine-tuned multilingual and English-based language models for the task of verifiable factual claim detection. Results of experiment 1 (translation to English) and experiment 2 (multi-lingual) are shown.

The test set for the verifiable factual claims detection task for 5 languages is almost balanced and contains 1826 and 1872 non-verifiable factual claims and verifiable factual claims respectively, hence it is proper to be tested for accuracy. Test subsets for singular languages are not balanced. The test set for harmful claims detection for 5 languages is strongly unbalanced and contains 3147 and 502 non-harmful claims and harmful claims respectively, and hence it is tested for f1-score as originally in the CLEF2022 paper. The details of test sets are presented in Appendix D.

The main metric that was taken into account was recall. Recall is used when the model's failure to detect the sought phenomenon (in our case, a specific post) results in significant negative effects. It was considered that it is a much worse case for fact-checkers to overlook a post with verifiable factual claims and harmful claims than to unnecessarily check a post that does not contain verifiable factual claims or harmful claims. The answer to the research question a) is included in Table 5, which presents that the use of multilingual models and training them on multilingual data in the verifiable factual claims detection task slightly worsens recall and at the same time improves accuracy. This conforms the [5] paper which reports that training xlm across multiple languages improves accuracy in downstream tasks. As shown in Table 5 the best results for recall were obtained for models based on the BERT-large-uncased (recall=0.7938) and the XLM-RoBERTa-large (recall=0.7724) architecture. Note that Table 5 compares results from both experiment 1 and experiment 2, because we are comparing setups (translated vs. multi-lingual) as well as model architectures. Including above results and the fact that multilingual models do not require translations, it was considered that they are more suitable to perform verifiable factual claims detection tasks than monolingual models trained on data translated

| Language of the test dataset | CLEF2022 best | XLM-RoBERTa-large with 10 epochs | XLM-RoBERTa-base | Multi-label XLM-RoBERTa-base |
|---|---|---|---|---|
| BG | 0.839 | 0.8176 | **0.8511** | 0.7964 |
| NL | **0.736** | 0.7239 | 0.7172 | 0.6966 |
| AR | 0.570 | **0.7861** | 0.7660 | 0.7676 |
| Exp. # | – | 2 | 2 | 5 |

**Table 6.** Accuracy comparison for detection of verifiable factual claims. Comparison between fine-tuned multilingual models (experiment 2), multi-label model (experiment 5) and results from the CLEF2022 (task 1B) paper [22]. Tests were conducted separately for three low-resource original languages from the CLEF2022 dataset.

into English. The XLM-RoBERTa models, unlike BERT-large-uncased, do not require English-only texts.

The answer to the research question b) is contained in tables Table 6 and Table 7, which showed that multilingual models perform well in the task of verifiable factual claims detection for low resource languages. The credibility of the presented models was verified by comparing with the results from the CLEF2022 [22] paper and with Nithiwat models (Table 7). In Table 6, results for three different languages are included, and the models were tested on the same testing datasets as the models from CLEF2022 (in terms of accuracy). Note that the table again combines results from multiple experiments (2 and 5): the idea is to compare both single- and multi-label models against the best results on CLEF2022. In comparison to the best models from CLEF2022, task 1B [22], the proposed models showed better accuracy in two cases (BG and AR), similar accuracy in one case (NL). The Multi-label XLM-RoBERTa-base model performs worse than the other models for BG, NL, CS and better for AR, SK, PL languages. The deterioration of the results should be explained by: a different training set (from collection 3); including data for harmful claims detection task; and also by due to that the training set included the MultiClaim dataset and the CLEF2021 dataset, i.e. datasets relating to topics other than Covid-19. Nevertheless, these results are still at the top of the list. The better results of the Multi-label XLM-RoBERTa-base model that occurred for SK and PL should be explained by the training set, which, unlike the Nithiwat models, also contained data in low-resource languages. There are discrepancies between the results for singular languages. In case of accuracy it can be even 16 points between the lowest for Dutch language result and the best for Bulgarian language result, and in case of recall it can be 5 points (between lowest for Czech language and best for Polish language). The explanation for this phenomenon is connected to differences between specific languages regarding the subsets of test data but also can be explained due to the XLM-RoBERTa architecture features. Based on [5] paper it is reported a strong dependence of the XLM-RoBERTa on the CC-100 training data corpus on which the architecture was trained from scratch.

| Language of the test dataset | Nithiwat mdeberta-v3-base claim-detection | Nithiwat xlm-roberta-base claim-detection | [Proposed] Multi-label XLM-RoBERTa-base |
|---|---|---|---|
| translations into SK | 0.8167 | 0.8378 | **0.8749** |
| translations into CS | 0.8165 | **0.8467** | 0.8348 |
| translations into PL | 0.8122 | 0.8071 | **0.8801** |

**Table 7.** Recall comparison between fine-tuned multi-label model (experiment 5) for verifiable factual claims detection and results based on Nithiwat models from Hugging Face [24] [23]. Tests were conducted for three low-resource languages SK, CS, PL.

| Model | Recall | Accuracy |
|---|---|---|
| Multi-label XLM-RoBERTa-base | 0.8024 | **0.7328** |
| Nithiwat mdeberta-v3-base claim-detection | 0.8120 | 0.7212 |
| Nithiwat xlm-roberta-base claim-detection | **0.8574** | 0.6898 |
| Alpaca-LoRA* | 0.5860 | 0.5408 |
| LLaMa-3.1-70b-instruct* | 0.4824 | 0.6733 |
| LLaMa-3.1-405b-instruct* | 0.7142 | 0.7066 |
| Qwen3.5-397b-a17b* | 0.8045 | 0.7271 |

**Table 8.** Task of verifiable factual claims detection, metric recall for the positive class and overall accuracy. Comparison between multi-label, multilingual language model (experiment 5) and external models (Nithiwat, Alpaca-LoRA, LLaMa-3.1-405b-instruct, LLaMa-3.1-70b-instruct, Qwen3.5-397b-a17b) [24] [23] [16] [8]. Tests done for 5 original languages from the CLEF2022 dataset; * tests done without data pre-processing.

For languages that had a large share in the CC-100 corpus the XLM-RoBERTa architecture shows much better results than for languages with a much smaller share in the corpus [5]. For Bulgarian language CC-100 corpus contains 57.5 GB of data and for Dutch 29.3 GB of data, for Polish and Czech 44.6 GB and 16.3 GB respectively. Based on proposed models it was reported, that the highest values were obtained for Bulgarian and Polish languages. This phenomenon is observed also for the final Multi-label XLM-RoBERTa-base model.

The answer to the research question c) is contained in Table 8 and Table 9, which show that the the Multi-label XLM-RoBERTa-base model does not achieve the best results for all the tested metrics, nevertheless, for the verifiable factual claims detection task, the model accuracy achieved the highest grade and the recall above 0.8 points, for the harmful claims detection task the f-1 score is lower than the best result. These results show that the Multi-label XLM-RoBERTa-base model performance does not differ significantly from the results of the best presented models. The obtained results are sufficiently high to use the model in hands-on NLP applications.

The answer to the research question d) is contained in Additional Analyses of multi-label XLM-RoBERTa-base model, which details are in Section 7. The main findings of the analysis is that for verifiable factual claims detection the model predictions are strongly connected to the sentence length. Longer sentences have significantly higher prediction score than the shorter sentences. The above finding was not observed in case of harmful claims detection.

| Model | f1-score | Test dataset |
|---|---|---|
| Multi-label XLM-RoBERTa-base | 0.3741 | 5 original languages |
| XLM-RoBERTa-base | 0.4032 | 5 original languages |
| Nithiwat mdeberta-v3-base claim-detection | 0.2431 | 5 original languages |
| Nithiwat xlm-roberta-base claim-detection | 0.2558 | 5 original languages |
| DistilBERT-base-uncased | 0.3587 | EN + translations into EN |
| Alpaca-LoRA* | 0.2108 | 5 original languages |
| LLaMa-3.1-70b-instruct* | 0.4139 | 5 original languages |
| LLaMa-3.1-405b-instruct* | 0.4539 | 5 original languages |
| Qwen3.5-397b-a17b* | **0.4664** | 5 original languages |

**Table 9.** Task of harmful claims detection, metric f1-score for the positive class. Comparison between fine-tuned models and external models (Nithiwat, Alpaca-LoRA, LLaMa-3.1-405b-instruct, LLaMa-3.1-70b-instruct, Qwen3.5-397b-a17b) [24] [23] [16] [8]. Results of experiment 3 and experiment 5. Tests done for 5 original languages and for translations to EN, test data from CLEF2022 [22]; * tests done without pre-processing.

## 7 Error Analyses

The answer to the research question d) is based on an error analysis for the multi-label XLM-RoBERTa-base model. The analysis revealed of 373 samples faultily classified as not verifiable factual claims (false negative) and 278 samples faultily classified as not harmful claims (false negative). The samples were analyzed in terms of sentence length, significant differences were detected in this respect, as the shortest sentence contained 27 characters and the longest sentence in this group contained 286 characters, this applies to a group of 373 samples. Similar differences occurred in the group of 278 samples related to harmful claims. For the analysis, the group of 373 samples was divided into 4 subsets, the first set contained sentences whose length was within the shortest 25% statistical length range, the second subset contained sentences whose length was between 25% and 50% of the statistical sentence length, the next subset concerned sentences whose length is in the range of 50% to 75% and the last set are sentences whose statistical length is the longest and ranges from 75% to 100% of the statistical sentence length, where 100% is the maximum length of the sentence. That is, in the range 0%-25% there were sentences with a minimum length of 26 characters and a maximum length of 95 characters; in the range of 25%-50% sentences longer than 95 characters and shorter than 145 characters; in the 50%-75% range, sentences longer than 145 characters and shorter than 223 characters, and in the 75%-100% allocation, sentences with more than 223 characters.

Analysis of the multi-label XLM-RoBERTa-base model predictions for verifiable factual claims detection revealed that there are significant differences in individual groups of sentences. Sentences with the longest length achieved significantly higher prediction scores than sentences with a much shorter length.

This leads to the conclusion that it can be assumed that the length of the sentence has a significant impact on the model prediction score and longer sentences are sentences that are more likely to contain verifiable factual claims. To support the findings additional analyses were conducted and recall versus the sentence length distribution was calculated, the results are presented in Appendix E of additional materials. The analyse confirms the findings, the highest recall was obtained for the group of the longest sentences and there is significant discrepancy (6 points) between the shortest and the longest group of sentences. Moreover the finding was compared to data from the literature, which suggests that longer sentences are more likely to contain truth than shorter sentences [18]. Therefore, shorter sentences more often contain fake news and longer ones contain truth. The hypothesis was proposed, that this similarity is not accidental and it is justified from the fact that when analyzing verifiable factual claims, in accordance with the above definition, the truth is examined and true sentences are usually longer. The detected relationship that longer sentences contain a more likely verifiable factual claim is, based on the proposed hypothesis, in accordance with information contained in the literature regarding the influence of sentence length on its truthfulness. To improve the model performance it is probably necessary to increase the number of samples (in the training set), which are relatively short sentences and they are containing verifiable factual claims also the possible way to improve the model performance can be to include the sentence length as additional (next to the text) feature during pre-training process. Both improvements applied together should make model more sentence length independent and more accurate in short sentence length detection containing verifiable factual claims, it will be our future research.

A similar analysis was carried out with regard to the detection of harmful claims, but no similar phenomenon was detected. Based on the analysis, it cannot be concluded that longer sentences are more likely to contain harmful claims. The obtained results show that the discrepancies in predictions for sentences of different lengths are relatively small and no clear trend in this respect was spotted. Fig.6 and Fig.7 in the Appendix E show respectively the probability score of the multi-label XLM-RoBERTa-base model for positive class of verifiable factual claim detection versus sentences length distribution and the probability score of the multi-label XLM-RoBERTa-base model for positive class of harmful claim detection versus sentences length distribution.

## 8 Conclusion

The paper has demonstrated the effectiveness of fine-tuned, pre-trained language models in accurately detecting both verifiable factual claims and harmful claims in posts, using a multi-label setup. It has also shown how the models perform at inference. Several models were trained and compared, starting with unilingual (EN) models, through multilingual models, and ending with multilingual, multi-label models.

Additionally, this study examined how multilingual models perform with low-resource languages such as Arabic, Bulgarian, Dutch, Slovak, Czech and Polish in the task of detecting verifiable factual claims. The presented results confirm that the fine-tuned models are credible and efficient for detecting both verifiable factual claims and harmful claims in posts. All reported results are based on evaluation of withheld test sets. The trained models can be successfully used as applicable tools to support manual fact-checking processes conducted by humans. Future work will focus on studying the impact of specific data on multi-label models.